\documentclass{article}
\usepackage{spconf,amsmath,graphicx}

\usepackage{graphicx}
\usepackage{amsmath}
\usepackage{dsfont}
\usepackage{amssymb}
\usepackage{booktabs}
\usepackage{subcaption}
\usepackage{tikz}
\usepackage{pgfplots}

\usepackage[pagebackref,breaklinks,colorlinks]{hyperref}

\title{
Pay Attention to Where You Looked\thanks{
This work was supported by STR (under a contract with DARPA) and Arete (with STTR research funding).\\
* Equal Contribution\\
Code: \href{https://github.com/alexberian/ViewAttention/}{https://github.com/alexberian/ViewAttention/}
}
}
\name{Alex Berian*, JhihYang Wu*, Daniel Brignac, Natnael Daba, Abhijit Mahalanobis}
\address{University of Arizona, ECE Dept.\\ Tucson, AZ, USA\\
\tt\small
[berian, jhihyangwu, dbrignac, ndaba, amahalan]@arizona.edu}

\begin{document}
%
\maketitle
\begin{abstract}
Novel view synthesis (NVS) has advanced with generative modeling, enabling photorealistic image generation. In few-shot NVS, where only a few input views are available, existing methods often assume equal importance for all input views relative to the target, leading to suboptimal results.

We address this limitation by introducing a camera-weighting mechanism that adjusts the importance of source views based on their relevance to the target. We propose two approaches: a deterministic weighting scheme leveraging geometric properties like Euclidean distance and angular differences, and a cross-attention-based learning scheme that optimizes view weighting. Additionally, models can be further trained with our camera-weighting scheme to refine their understanding of view relevance and enhance synthesis quality. This mechanism is adaptable and can be integrated into various NVS algorithms, improving their ability to synthesize high-quality novel views. Our results demonstrate that adaptive view weighting enhances accuracy and realism, offering a promising direction for improving NVS.
\end{abstract}


\section{Introduction}
\label{sec:intro}

Novel view synthesis (NVS) has seen a surge of interest in this current age of generative modeling. While previous NVS method achieved significant results \cite{nerf}, the introduction of diffusion models \cite{originaldiffusion,mvdream} has pushed the field of NVS closer to generating photorealistic images indistinguishable from real images. Due to diffusion's generative power, many recent works \cite{genvs, reconfusion, cmd} leverage a diffusion model's capabilities when rendering the final image. 

In novel view synthesis, we are given a set of input images with respective camera poses and tasked with generating a new, unique target image from an unseen target camera pose. Neural radiance field (NeRF) style methods \cite{nerf, mipnerf, zipnerf} use a dense set of input images to train a neural network to predict the target view's color and density and proceed to volume rendering to obtain the final output image.

More recent NVS methods assume we are given the target camera pose of the image we wish to generate \cite{pixelnerf, genvs}. The target camera pose is then used to cast rays in 3D space. These rays are then sampled to obtain latent vectors in the 3D world that can be transformed into each of the input image's perspectives and used as input to a traditional NeRF model to obtain a rendered image. 

\begin{figure}
    \centering
    \begin{subfigure}[b]{0.225\textwidth}
        \centering
        \includegraphics[trim=0 315 487 0, clip, width=\textwidth]{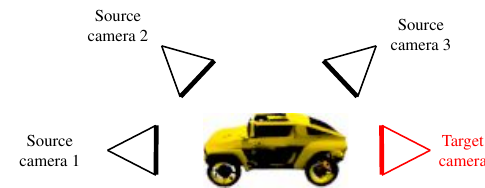}
        \caption{Without weighting}
        \label{fig:subfig-a}
    \end{subfigure}
    \hfill
    \begin{subfigure}[b]{0.225\textwidth}
        \centering
        \includegraphics[trim=0 165 487 154, clip, width=\textwidth]{figures/intro-fig.pdf}
        \caption{With weighting}
        \label{fig:subfig-b}
    \end{subfigure}
    \caption{Example of the source camera weighting problem. (a) shows equal weighting of camera views even though some views capture non-useful information, and (b) shows prioritized weighting on the darker source camera as it is closest to the target view.}
    \label{fig:main-fig}
\end{figure}

While the above methods make use of target camera information, they treat the importance of each source view relative to the target view as equal. Naturally, we would not expect this to be the case as some source views may contain little to no information pertaining to the queried target view. For example, if the target view queries the dorsal side of an object and we have 3 input views, 2 corresponding to frontal views and 1 corresponding to a side dorsal view, then we would expect the side dorsal input view to have a higher weighting when performing the rendering process. Figure \ref{fig:main-fig} depicts this example.

In this work, we seek to solve this problem of source camera weighting with respect to the target view. We propose two schemes to accomplish this task: a deterministic weighting scheme and a cross-attention-based learning scheme. In the deterministic scheme, we experiment with various forms of deterministic weighting such as relative Euclidean distance from source views to the target view and error weighting based on differences in angles and camera distances from source views to the target view. In the learned scheme, we use the concept of cross-attention \cite{transformer} for the model to learn which source views should attend to the target view more. Our experiments demonstrate that camera weighting outperforms the traditional scenario where all input poses are weighted equally.

\section{Overview}
\subsection{Novel View Synthesis}
NVS is the problem of inferring a target image $x_t$ from $S$ source images $\{x_{si}\}_{i=1}^S$\footnote{For brevity, we denote $\{x_{si}\}_{i=1}^S = \{x_{s1}, x_{s2}, x_{s3}, \dots, x_{sS}\}$}. The target images and source images have associated camera pose matrices $P_t\in\mathds{R}^{(4,4)}$ and $\{P_{si}\}_{i=1}^S$ respectively. Few-shot NVS is when $S$ is small (e.g. 5 or less). We seek to create a NVS model $N_\theta$ with parameters $\theta$ that predicts novel views
\begin{equation}
    \hat{x}_t = \ N_\theta(P_t,\{x_{si}\}_{i=1}^S,\{P_{si}\}_{i=1}^S)
    .
    \label{eq:nvs}
\end{equation}

\subsection{GeNVS}
GeNVS \cite{genvs} encodes each source image into feature volumes $W_i \in \mathds{R}^{128,128,64,16}$ (point clouds of 16-dimensional features) using a modified DeepLabV3+ \cite{deeplabv3p,segmentationgithub} segmentation model. The volumes are oriented as truncated rectangular pyramids with respect to their source camera poses $\{P_{si}\}_{i=1}^S$ and a dataset-specific $z_{near}$ to $z_{far}$. A 16-channel feature image $F \in \mathds{R}^{64,64,16}$ of the volumes is rendered via volume rendering from a target camera pose $P_t$. Stratified sampling is used to select points along each ray. A point $r$ is trilinearly interpolated in each feature volume to get latent vectors $\mathbf{l}_i(r) = W_i(r) \in \mathds{R}^{16}$. The latent vectors are averaged together, then passed through a MLP $f$ to get 
a 16-channel color $c$ and density $\sigma$.
\begin{equation}
    c,\sigma = f\Bigg(\sum_{i=1}^S\mathbf{l}_i(r)\Bigg)
    .
    \label{eq:genvs_avg}
\end{equation}
Once the color and density of every point along a ray are predicted, volume rendering is used to predict the 16-channel pixel value.

The 16-channel feature image is used as conditioning to a denoising diffusion \cite{edm} U-Net $U$ by concatenating it with 3-channel noise (or noisy target image). The denoising U-Net $U$ outputs 3 channels. When sampling new images with multiple denoising steps, the feature image is skip-concatenated to the U-Net's $U$ input. The final output target image $x_t$ is the output from the last denoising step. 

\subsection{PixelNeRF}
PixelNeRF, like GeNVS, predicts latent vectors $\mathbf{l}_i(r)$ for points $r$ along a ray for each source image $x_i$ and uses a neural network to predict color $c$ and density $\sigma$ for volume rendering. 

PixelNeRF encodes each input image to a grid of latent vectors $G_i \in \mathds{R}^{64,64,128}$. A ray point $r$ is projected onto $S$ source image $\{x_{si}\}_{i=1}^{S}$ plane then grid vector $G_i(r) \in \mathds{R}^{128}$ is bilinearly interpolated. Each grid vector is concatenated with a positional encoding of the point's $r$ position in space, and the view direction with respect to the source pose $P_{si}$. This resulting vector is passed through a few fully connected ResNet \cite{resnet} blocks to produce a latent vector $\mathbf{l}_i(r)$. As in equation (\ref{eq:genvs_avg}) The latent vectors $\{\mathbf{l}_i(r)\}_{i=1}^S$ are averaged followed by a few more fully connected ResNet blocks $f$ to produce the final 3-channel color $c$ and density $\sigma$ of the point $r$. The rendered rays give the pixel values of the target image $x_t$.

\section{Problem Statement}
\label{sec:problem}
PixelNeRF and GeNVS's averaging of latent vectors does not take into account the relationship between the source poses $\{P_{si}\}_{i=1}^S$ and the target pose $P_t$. (\ref{eq:genvs_avg}) can be generalized as a weighted average
\begin{equation}
    c,\sigma = f\Big(\mathbf{l}(r)\mathbf{w}\Big) = f\Bigg(\sum_{i=1}^S\mathbf{l}_i(r)\mathbf{w}_i\Bigg)
    ,
    \label{eq:weighted-avg}
\end{equation}
where $\mathbf{l}(r) \in \mathds{R}^{L,S}$ is a matrix of latent vectors and $\mathbf{w} \in \mathds{R}^{S}$ is a vector of weights per source image. The weight's vector is for weighted averaging, therefore it must satisfy the constraint.
\begin{equation}
    \sum_{i=1}^{S}\mathbf{w}_i = 1
    .
    \label{eq:constraint}
\end{equation}
In the averaging case of (\ref{eq:genvs_avg}), all weights are equal $\mathbf{w}=[\frac{1}{S},\frac{1}{S},...]^T$.

We seek to devise a camera weighting function
\begin{equation}
    \mathbf{w} = C(P_t,\{P_{si}\}_{i=1}^S)
\end{equation}
that utilizes the source poses and the target pose.

\section{Method}
\label{sec:method}
We propose two types of camera weighting methods: deterministic weighting, and attention-based weighting. Deterministic weighting is directly calculated from source and target poses, whereas attention-based weighting utilizes neural attention mechanisms \cite{transformer}. The proposed weighting methods can simply substitute the averaging step without the need to retrain the entire NVS model. The attention-based approaches do need to be trained, but the rest of the NVS model's parameters may be fixed during training.

\subsection{Deterministic Weighting}
We experimented with various deterministic weighting approaches. Each approach calculates intermediate weights that do not satisfy the constraint in (\ref{eq:constraint}). The actual weights $\mathbf{w}$ are derived from intermediate weights $\mathbf{w}'$ with normalization:
\begin{equation}
    \mathbf{w} = \frac{\mathbf{w}'}{\sum_{i=1}^{S} \mathbf{w}_i'}
    .
    \label{eq:weight-fixing}
\end{equation}

\emph{L1 Norm Weighting} and \emph{Frobenius Norm Weighting} are given by
\begin{equation}
    \mathbf{w}_i' = \frac{1}{ \varepsilon + Norm}
    ,
    \label{eq:l1}
\end{equation}
where $\varepsilon=10^{-6}$ and $Norm$ is the L1 norm or Frobenius norm between the source pose matrix $P_{si}$ and $P_t$.

\subsubsection{Distance Gaussian Kernel Weighting}
Distance Gaussian Kernel weighting applies the Gaussian kernel to the distance between the target camera center $\mathbf{c}_t \in \mathds{R}^{3}$ and the source camera centers $\{\mathbf{c}_{si}\}_{i=1}^{S}$
\begin{equation}
    \mathbf{w}_i' = e^{-\beta\|\mathbf{c}_t-\mathbf{c}_{si}\|^2}
    ,
    \label{eq:gauss-kernel}
\end{equation}
where $\|\cdot\|$ is vector magnitude, and $\beta$ is a hyper-parameter to control how much to prioritize closer source cameras over further ones.

\subsubsection{Error Weighting}
Error weighting is a combination of camera distance and view-angle difference. It is given by 
\begin{equation}
    \mathbf{w}_i' = \frac{1}{\varepsilon + 
            \alpha \frac{\theta_i}{\pi} + 
            (1-\alpha)\frac{\|\mathbf{c}_t-\mathbf{c}_{si}\|}
                           {\max_{k=1}^S\|\mathbf{c}_t-\mathbf{c}_{sk}\|}
            }
    ,
    \label{eq:cam-error}
\end{equation}
where $\theta_i \in \mathds{R}$ is the angle between the principle view axis of the target view and the $i^{th}$ source view, and $\alpha$ is a hyper-parameter to tune how much to prioritize angle error over camera center error. When $\alpha=1$, only the angle error is used, and when $\alpha=0$, only camera distance is used.

\subsection{Attention-Based Weighting}
Attention-based weighting begins with embedding target/source poses. The camera embeddings are passed into an attention mechanism \cite{transformer} to get weights $\mathbf{w}$ that satisfy the constraint of \ref{eq:constraint}.

\subsubsection{Pose Embedding}
Camera pose embedding takes a source or target camera pose matrix $P \in \mathds{R}^{4,4}$, and processes it into a vector. We experimented with a few camera embeddings, but the two that performed the best are shown here.

\begin{figure}
  \centering
   \includegraphics[trim=0 186 598 0, clip,angle=0,width=90pt]{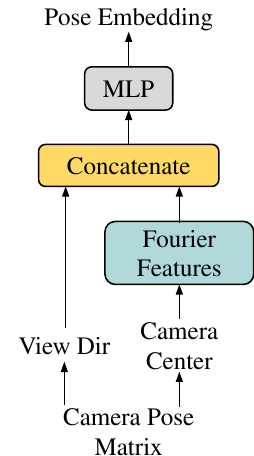}
   \caption{Process for converting a camera pose matrix to a pose embedding vector.}
   \label{fig:pose-embed}
\end{figure}

The first embedding approach, shown in Fig. \ref{fig:pose-embed}, extracts the camera center $\mathbf{c}$ and the view direction $\mathbf{v} \in \mathds{R}^{3}$ from the pose matrix $P$. We use positional encoding \cite{positionalencoding} (Fourier features) on the camera center, concatenate that with the view direction unit vector, and then an tiny MLP with ReLU activations.

The second approach is to flatten the camera pose matrix $P$, followed by 2 linear layers with ReLU activations. In practice, we found that the embedding in Fig. \ref{fig:pose-embed} achieved the best overall performance.

\subsubsection{Cross-Attention Weighting}
\label{sec:cross-attention}

\begin{figure}
  \centering
   \includegraphics[trim=0 209 548 0, clip,angle=0,width=130pt]{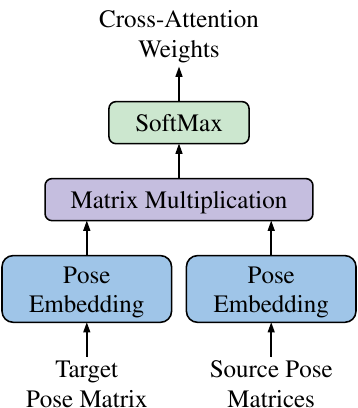}
   \caption{Process of the cross-attention weighting algorithm.}
   \label{fig:cross-attention}
\end{figure}

The idea behind cross-attention weighting (CAW), shown in Fig. \ref{fig:cross-attention}, is to use a learned manner of relating the target pose to all the source poses. We embed the target poses into an embedding vector $\mathbf{E}_t  \in \mathds{R}^{1,A}$ where $A$ is the attention vector length, and the source pose embeddings are placed into a matrix $\mathbf{E}_s \in \mathds{R}^{S,A}$. We correlate the target embedding with the source embeddings with matrix multiplication to get $S$ intermediate weight values, then use softmax to satisfy the constraint in (\ref{eq:constraint}). Formally, 
\begin{equation}
    \mathbf{w}_{CAW} = softmax\bigg( \frac{\mathbf{E}_t \mathbf{E}_s\textsuperscript{T}}{\sqrt{A}} \bigg)
    .
    \label{eq:cross-attention-weighting}
\end{equation}
As done in \cite{transformer} we divide by $\sqrt{A}$ to ensure gradients through softmax don't vanish as the attention dimension $A$ becomes large.

\subsection{Training}
We begin with pre-trained models that utilize baseline (mean) for source camera weighting. Deterministic approaches do not require training.

Attention-based weighting methods do need to be trained. We freeze the model, but not the camera weighting module, and train with the same learn rate and loss with which the full model was trained.

\subsection{Implementation Details}
We perform all experiments with PyTorch \cite{pytorch}. We cloned the PixelNeRF GitHub repository and made our modifications. For GeNVS, we re-implemented it by building on top of EDM's\cite{edm} GitHub repository. We use the provided pre-trained weights for PixelNeRF and train our GeNVS models from scratch until they match the published performance. We use 25 denoising steps through all our experiments involving GeNVS.

In pose embedding, we use the positional encoding module from PixelNeRF with $6$ frequencies, including the input, and a frequency factor of $1.5$. The MLP is composed of 3 linear layers and ReLU activations between them. We use a hidden dimension of $64$ and a output attention dimension of $A=128$.

\section{Experiments}
\label{sec:experiments}

In this section we present our experiments on the SRN Cars and SRN Multi-Chairs dataset \cite{srn}. For brevity, we only show numerical experimental results for SRN Cars. For experiments on additional datasets, we refer the reader to the appendix.

\subsection{Weighting Methods}
We compare various camera weighting methods on randomly selected source and target views in Table \ref{tab:comparison}. We observe that error weighting achieves the best performance with the SRN Cars \cite{srn} dataset both for PixelNeRF and GeNVS. L1 weighting and cross-attention weighting performed the best for GeNVS and distance Gaussian for PixelNeRF on the SRN multi-chairs dataset\cite{pixelnerf, srn}.

\begin{table*}[htbp]
\centering
\small 
\setlength{\tabcolsep}{4pt} 
\begin{tabular}{cccccc|ccccc}
\toprule
\multicolumn{6}{c|}{\textbf{PixelNeRF}} & \multicolumn{5}{c}{\textbf{GeNVS}} \\
\hline
Method & \begin{tabular}[c]{@{}c@{}}FID↓\end{tabular} & \begin{tabular}[c]{@{}c@{}}LPIPS↓\end{tabular} & \begin{tabular}[c]{@{}c@{}}DISTS↓\end{tabular} & \begin{tabular}[c]{@{}c@{}}PSNR↑\end{tabular} & \begin{tabular}[c]{@{}c@{}}SSIM↑\end{tabular} &
\begin{tabular}[c]{@{}c@{}}FID↓\end{tabular} &
\begin{tabular}[c]{@{}c@{}}LPIPS↓\end{tabular} &
\begin{tabular}[c]{@{}c@{}}DISTS↓\end{tabular} &
\begin{tabular}[c]{@{}c@{}}PSNR↑\end{tabular} &
\begin{tabular}[c]{@{}c@{}}SSIM↑\end{tabular}\\
\hline
Baseline (mean)            & 29.436          & 0.0857          & 0.155          & 26.961          & 0.949          & 6.294           & 0.0574          & 0.113          & 24.958          & 0.939          \\
Dist. Gauss $\beta=1$      & 23.508          & 0.0753          & 0.146          & 27.249          & 0.950          & 5.622           & 0.0523          & 0.105          & 25.345          & 0.942          \\
Dist. Gauss $\beta=0.3$    & 25.230          & 0.0775          & 0.147          & 27.338          & 0.952          & 5.973           & 0.0543          & 0.108          & 25.288          & 0.942          \\
L1 Weighting               & 24.113          & 0.0737          & 0.144          & 27.633          & 0.953          & 5.483           & 0.0501          & 0.103          & 25.724          & 0.946          \\
F-norm Weighting           & 23.927          & 0.0738          & 0.144          & 27.660          & 0.954          & 5.541           & 0.0505          & 0.104          & 25.700          & 0.945          \\
Err Weighting $\alpha=0.5$ & 23.785          & 0.0735          & 0.144          & 27.634          & 0.953          & 5.559           & 0.0506          & 0.104          & 25.687          & 0.945          \\
Err Weighting $\alpha=1.0$ & \textbf{23.500} & \textbf{0.0724} & \textbf{0.143} & \textbf{27.707} & \textbf{0.954} & \textbf{5.417}  & \textbf{0.0493} & \textbf{0.103} & \textbf{25.767} & \textbf{0.946} \\
Cross-Attention Weighting & N/A             & N/A             & N/A            & N/A             & N/A            & 5.727           & 0.0521          & 0.106          & 25.430          & 0.944          \\
\hline
\end{tabular}
\caption{\textbf{PixelNeRF and GeNVS on SRN cars.} We show how weighting algorithms can improve the performance of PixelNeRF and GeNVS by just weighting certain views more than others.}
\label{tab:comparison}
\end{table*}

\subsection{Close Input Views}
Our methods particularly excel when one of the input views is close ($<10^\circ$) to the target view. By weighting close input views more, noise from far views don't contribute as much to the combined latent used to produce the final rendering. We show this in Table \ref{tab:close}.

\begin{table*}[htbp]
\centering
\small 
\setlength{\tabcolsep}{4pt} 
\begin{tabular}{cccccc|ccccc}
\toprule
\multicolumn{6}{c|}{\textbf{PixelNeRF}} & \multicolumn{5}{c}{\textbf{GeNVS}} \\
\hline
Method & \begin{tabular}[c]{@{}c@{}}FID↓\end{tabular} & \begin{tabular}[c]{@{}c@{}}LPIPS↓\end{tabular} & \begin{tabular}[c]{@{}c@{}}DISTS↓\end{tabular} & \begin{tabular}[c]{@{}c@{}}PSNR↑\end{tabular} & \begin{tabular}[c]{@{}c@{}}SSIM↑\end{tabular} &
\begin{tabular}[c]{@{}c@{}}FID↓\end{tabular} &
\begin{tabular}[c]{@{}c@{}}LPIPS↓\end{tabular} &
\begin{tabular}[c]{@{}c@{}}DISTS↓\end{tabular} &
\begin{tabular}[c]{@{}c@{}}PSNR↑\end{tabular} &
\begin{tabular}[c]{@{}c@{}}SSIM↑\end{tabular}\\
\hline
\textbf{Random Input Views}\\
Baseline (mean)            &   83.259          & 0.139           & 0.204          & 23.053          & 0.905          &   145.069         & 0.440           & 0.374          & 12.645          & 0.747          \\
Err Weighting $\alpha=1.0$ &   71.930          & 0.125           & 0.192          & 23.433          & 0.910          &   96.353          & 0.392           & 0.329          & 13.664          & 0.754          \\
\hline
\textbf{1 Close Input View}\\
Baseline (mean)            &   54.699          & 0.0896          & 0.167          & 25.634          & 0.941          &   139.511         & 0.391           & 0.349          & 13.045          & 0.763          \\
Err Weighting $\alpha=1.0$ &   \textbf{43.219} & \textbf{0.0622} & \textbf{0.138} & \textbf{26.978} & \textbf{0.954} &   \textbf{35.208} & \textbf{0.124}  & \textbf{0.155} & \textbf{19.035} & \textbf{0.853} \\
\hline
\end{tabular}
\caption{\textbf{Close input views.} We show how weighting methods are especially important when an input view is close ($<10^\circ$) to the target view. For each scene tested, we randomly choose 5 input views and 1 random target view. For 1 close input view, we repeatedly sample input views until at least one of them is within $10^\circ$ of target view.}
\label{tab:close}
\end{table*}

\subsection{Number of Input Views}

\begin{figure}
    \centering
    \scalebox{0.7}{
        \begin{tikzpicture}
            \begin{axis}[
                xlabel={Number of Input Views},
                ylabel={PSNR},
                xmin=2, xmax=32,
                ymin=22, ymax=32,
                xtick={2, 8, 16, 32},
                ytick={22, 24, 26, 28, 30, 32},
                legend pos=north west,
                ymajorgrids=true,
                grid style=dashed,
            ]

                \addplot[
                    color=blue,
                    mark=square,
                    ]
                    coordinates {
                        (2, 24.56117031)
                        (8, 26.83243468)
                        (16, 27.22434984)
                        (32, 27.60971083)
                    };
        
                \addplot[
                    color=red,
                    mark=triangle,
                    ]
                    coordinates {
                        (2, 24.48693763)
                        (8, 27.65772429)
                        (16, 28.81414195)
                        (32, 29.68951503)
                    };
    
                \legend{Baseline (mean),Err Weighting $\alpha=1.0$}

            \end{axis}
        \end{tikzpicture}
    }
    
    \caption{\textbf{Weighting sustains growth on PixelNeRF.} While baseline (mean) performance plateaus with increasing input views, error weighting sustains the growth. Each method number of input views pair is evaluated on the entire SRN Cars test distribution with 3 random target views for each scene tested.}
    
    \label{fig:num_input_views_pixelnerf}
    
\end{figure}
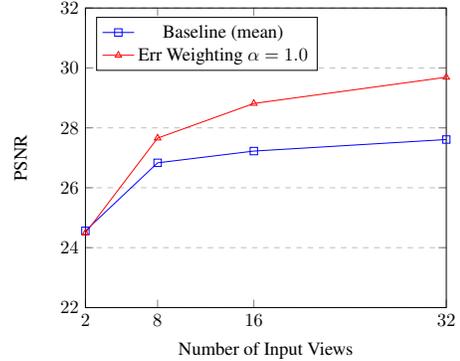

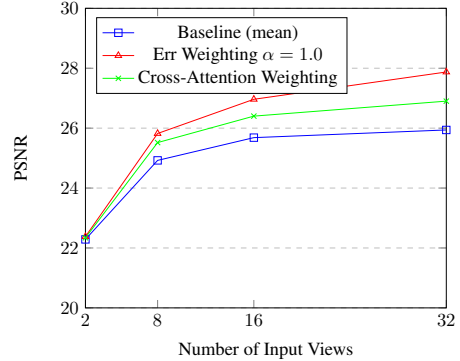
\begin{figure}
    \centering
    \scalebox{0.7}{
\begin{tikzpicture}
\begin{axis}[
    xlabel={Number of Input Views},
    ylabel={PSNR},
    xmin=2, xmax=32,
    ymin=20, ymax=30,
    xtick={2, 8, 16, 32},
    ytick={20, 22, 24, 26, 28, 30},
    legend pos=north west,
    ymajorgrids=true,
    grid style=dashed,
]

\addplot[
    color=blue,
    mark=square,
    ]
    coordinates {
    (2, 22.2861642)
    (8, 24.92349848)
    (16, 25.68310227)
    (32, 25.9394996)
    };
    
\addplot[
    color=red,
    mark=triangle,
    ]
    coordinates {
    (2, 22.38102128)
    (8, 25.82154479)
    (16, 26.96323635)
    (32, 27.87426729)
    };
    
\addplot[
    color=green,
    mark=x,
    ]
    coordinates {
    (2, 22.34097907)
    (8, 25.51791793)
    (16, 26.40143518)
    (32, 26.90147484)
    };

\legend{Baseline (mean),Err Weighting $\alpha=1.0$, Cross-Attention Weighting}

\end{axis}
\end{tikzpicture}
}
\caption{\textbf{Weighting sustains growth on GeNVS.} While baseline (mean) performance plateaus with increasing input views, error weighting and cross-attention weighting sustains the growth. Evaluation setup is the same as the one used in Figure \ref{fig:num_input_views_pixelnerf}.}
\label{fig:num_input_views_genvs}
\end{figure}

While baseline mean performance plateaus when the number of input views increases, our weighting methods sustains the growth by choosing views that are important and rejecting noise from other views that are not important. This can be observed Figures \ref{fig:num_input_views_pixelnerf} and \ref{fig:num_input_views_genvs}.

\subsection{Image comparisons}
\begin{figure}[p]
    \centering
    \includegraphics[trim=0 545 0 50, clip, width=170pt]{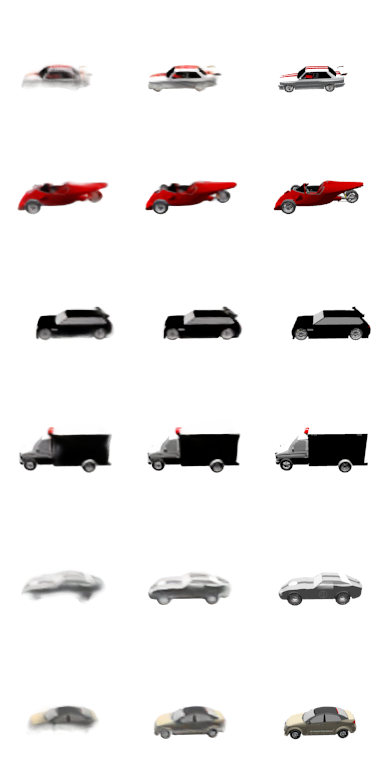}
    \caption{\textbf{Baseline (mean) versus Error Weighting using PixelNeRF on SRN cars.} Comparison of images generated by PixelNeRF when using weighting methods. Columns are baseline (mean), error weighting $\alpha=1.0$, and ground truth, respectively.}
    \label{fig:pixel_cars}
\end{figure}

\begin{figure}[p]
    \centering
    \includegraphics[trim=0 545 0 50, clip, width=170pt]{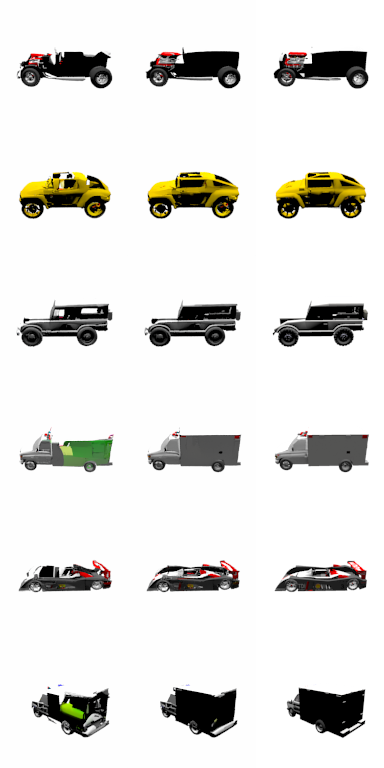}
    \caption{\textbf{Baseline (mean) versus Error Weighting using GeNVS on SRN cars.} Comparison of images generated by GeNVS when using weighting methods. Columns are baseline (mean), error weighting $\alpha=1.0$, and ground truth, respectively.}
    \label{fig:genvs_cars}
\end{figure}

We compare images generated by PixelNeRF and GeNVS using baseline (mean) and an improved weighting method. The images generated with improved weighting method in PixelNeRF are significantly sharper and more detailed. Similarly, GeNVS with improved weighting methods generates images that more closely resemble the ground truth and reduces anomalies by the diffusion model. We show examples in Figures \ref{fig:pixel_cars}, \ref{fig:genvs_cars}. We refer the reader to the appendix for additional image comparisons.

\section{Conclusion}
\label{sec:conclusion}

In this work, we addressed the limitations of current few-shot novel view synthesis (NVS) methods by introducing a camera-weighting mechanism to better account for the varying importance of input views in relation to the target view. We proposed both deterministic weighting scheme and a cross-attention-based learning approach. Our experiments, conducted on PixelNeRF and GeNVS, demonstrate that the weighting schemes outperform averaging, especially when one of the input views is close to the target view. Additionally, we show that as the number of input views increases, our camera-weighting methods more effectively utilize information from the input views. These results highlight the effectiveness of our approach in improving few-shot NVS quality and pave the way for more robust solutions in NVS.

\bibliographystyle{IEEEbib}
\bibliography{strings,refs}

\end{document}